\documentclass[a4paper]{article}

\usepackage[final]{nips_2017}

\usepackage[utf8x]{inputenc}
\usepackage[T1]{fontenc}

\usepackage{amsmath} 
\usepackage{graphicx}
\usepackage{algorithm}
\usepackage[noend]{algpseudocode}
\usepackage{booktabs} 

\usepackage{titlesec}
\titlelabel{\thetitle.\quad}

\makeatletter
\def\BState{\State\hskip-\ALG@thistlm}
\makeatother
\usepackage[draft]{hyperref}

\usepackage{natbib}

\usepackage[inline]{enumitem}

\usepackage{tikz}
\usetikzlibrary{shapes.geometric, arrows, fit}
\tikzstyle{io} = [trapezium, trapezium left angle=70, trapezium right angle=120, minimum width=2cm, minimum height=1cm, text centered, draw=black, fill={rgb:red,1;green,2;blue,5;white,6}, trapezium stretches=true]
\tikzstyle{process} = [rectangle, minimum width=2cm, minimum height=1cm, text centered, draw=black, fill={rgb:red,1;green,2;blue,5;white,20}]
\tikzstyle{decision} = [diamond, minimum width=3cm, minimum height=1cm, text centered, draw=black, fill={rgb:red,1;green,2;blue,5;white,1}]
\tikzstyle{arrow} = [thick,->,>=stealth]

\usepackage{xspace}

\newcommand{\ie}{i.e.,\xspace}

\newcommand{\Xf}{\textbf{X}}
\newcommand{\Pa}{\text{Pa}}

\usepackage[colorinlistoftodos]{todonotes}

\title{A Universal Marginalizer for Amortized Inference in Generative Models}

\author{Laura Douglas\thanks{The authors equally contributed to the manuscript.}\ \ \thanks{babylon, London, UK. Email for correspondence: \url{saurabh.johri@babylonhealth.com}}\ ,\ \  Iliyan Zarov\footnotemark[1]\ \ \footnotemark[2]\ ,\ \ Konstantinos Gourgoulias\footnotemark[2]\ , \ \ Chris Lucas\footnotemark[2]\ ,\\
\textbf{Chris Hart\footnotemark[2]\ ,\ \ Adam Baker\footnotemark[2]\ , \ \ Maneesh Sahani\thanks{Gatsby Computational Neuroscience Unit, University College London.}\ , \ \ Yura Perov\footnotemark[2]\ , \ \ Saurabh Johri\footnotemark[2]}}

\def\Pa{\mathrm{Pa}}

\def\XX{\mathbf{X}}
\def\XO{\mathbf{X}_{\mathcal{O}}}
\def\xO{\mathbf{x}_{\mathcal{O}}}
\def\xxO{\mathbf{\tilde x}_{\mathcal{O}}}
\def\xxS{\mathbf{\tilde x}_{\mathcal{S}}}
\def\XU{\mathbf{X}_{\mathcal{U}}}
\def\xU{\mathbf{x}_{\mathcal{U}}}

\def\xS{\mathbf{x}_{\mathcal{S}}}

\def\xxSuO{\mathbf{\tilde x}_{\mathcal{S}\cup\mathcal{O}}}

\def\xSnPai {\mathbf{x}_{\mathcal{S}\cap\Pa(X_i)}}

\def\UM{\mathrm{UM}}

\begin{document}
\maketitle
\begin{abstract}
We consider the problem of inference in a causal generative model
where the set of available observations differs between data
instances.  We show how combining samples drawn from the graphical model with an
appropriate masking function makes it possible to train a single neural
network to approximate all the corresponding conditional marginal
distributions and thus amortize the cost of inference.  We further demonstrate that the efficiency of importance sampling may be improved by basing proposals on the output of the neural network. We also outline how the same network can be used to generate samples from an approximate joint posterior via a chain decomposition
of the graph.
\end{abstract}

\section{Introduction}
Graphical models provide a natural framework for expressing probabilistic relationships between random variables, and are widely used to facilitate reasoning and decision-making. Bayesian networks (BN), a directed form of probabilistic graphical model (PGM), have been used extensively in medicine to capture the causal relationships between medical entities such as diseases and symptoms, and through inference, enable diagnosis of unobserved disease states. In sensitive domains such as health care, the penalty for errors in inference is potentially severe. This risk can be mitigated by increasing the complexity of the model of the underlying process; however, this will also increase the cost of inference, limiting the feasibility of most algorithms.

In complex models, exact inference is often computationally intractable. We therefore must resort to approximate  methods such as variational inference~\citep{wainwright2008graphical} and Monte Carlo methods, e.g.,  importance sampling~\citep{cheng2000ais,neal2001annealed}. Variational inference methods are fast but inexact; Monte Carlo inference is asymptotically exact, but can be slow. For this reason, we focus on Amortized Inference, techniques which speed up sampling by allowing us to ``flexibly reuse inferences so as to answer a variety of related queries''~\citep{gershman2014amortized}.
Amortized inference has been popular for Sequential Monte Carlo and has been used to learn in advance either parameters~\citep{gu2015neural} or a discriminative model which provides conditional density estimates~\citep{Morris:2001:RNA:2074022.2074068, paige2016inference}. These conditional density estimates can be used as proposals for importance sampling (see Appendix~\ref{sec:app:impo-sampling}), an approach also explored in~\citep{le2017using}, using a fixed sequential density estimator MADE~\citep{germain2015made}. We propose a related technique with a more general density estimator to allow arbitrary evidence.

\noindent
{\bf Notation:}
Consider the set of random variables, $\Xf=\{X_1,\ldots, X_N\}$. A BN is a combination of a Directed Acyclic Graph (DAG), with $X_i$ as nodes, and a joint distribution $P$ of the $X_i$. The distribution $P$ factorizes according to the structure of the DAG with $P(X_i|\Pa(X_i))$ being the conditional distribution of $X_i$ given its parents, $\Pa(X_i)$. We denote by $\XO$ a set of instantiated nodes, $\mathbf{X_\mathcal{O}}\subset \Xf$, and their instantiations $\xO$. To conduct Bayesian inference when provided with a set of unobserved nodes, $\mathbf{X_\mathcal{U}}\subset \Xf\setminus\mathbf{X_\mathcal{O}}$, we need to compute the posterior marginal, $ P(\mathbf{X_\mathcal{U}}|\mathbf{X_\mathcal{O}})$.

\section{Universal Marginalizer}
\label{section:um}

We consider a function approximator, such as a neural network (NN), trained to return the marginal posterior distributions $P(X_i| \XO = \xO)$ for each node $X_i \in \XX$ given an instantiation $\xO$ of \emph{any} set of observations $\XO \subset \XX$. We write $\xxO$ for an encoding of the instantiation that specifies both which nodes are observed, and what their values are -- some suitable encodings are discussed in Section~\ref{sec:methods} below.  Then, for binary $X_i$, the desired network maps $\xxO$ to a vector in $\{0,1\}^N$ representing the probabilities $p_i:=P(X_i= 1 | \XO = \xO)$:
\begin{align}
	\label{eq:umy}
  \mathbf{Y}=\UM(\xxO) \approx (p_1,\ldots, p_N).
\end{align}
To approximate any possible posterior marginal distribution (\ie given any possible set of evidence $\mathbf{X_\mathcal{O}}$), only one model is needed. Due to this we describe this discriminative model as a \textit{Universal Marginalizer} (UM).  The existence of such a network is a direct consequence of the universal function approximation theorem (UFAT)~\citep{hornik}. This is simply illustrated by considering marginalization in a BN as a function and that, by UFAT, any continuous function can be arbitrarily approximated by a NN. 

\subsection{Training a UM}

\noindent {\bf 1.\ Generating Data.} The UM can be trained off-line by generating unbiased samples from the BN using ancestral sampling \cite[Algorithm 12.2]{koller2009probabilistic}. Each sample is a binary vector which are the values the classifier will learn to predict.

\noindent {\bf 2.\ Masking.} For the purpose of prediction, a subset of the nodes in the sample must be hidden, or  \textit{masked}. This masking can be deterministic, \ie always masking specific nodes, or probabilistic over nodes. We choose to probabilistically mask a sample in an unbiased way by defining a masking probability, $p \sim U[0,1]$, which is applied to each node. This will create a dataset whose number of observed nodes is uniformly distributed. There is some analogy here to dropout in the input layer and so could work well as a regularizer, independently of this problem \citep{JMLR:v15:srivastava14a}.

\noindent {\bf 3.\ Representation of the Unobserved/Masked Nodes.} Masked nodes are created for the purpose of mimicking unobserved nodes and so the representation of masked nodes must be consistent with the unobserved nodes at the time of inference. Different representations will be investigated and tested in Section~\ref{sec:methods}.

\noindent {\bf 4.\ Training a neural net with Cross Entropy Loss.} By~\citep{saerens2002any}, the output of a NN with any reasonable loss function can be mapped to a probability estimate, however the cross entropy loss is the most obvious choice as the output is exactly the predicted probability distribution. We train the network using a binary cross entropy loss function in a multi-label classification setting to predict the state of all observed and unobserved nodes.

\noindent {\bf 5.\ Outputs: Posterior Marginals.} The desired posterior marginals are the output of the sigmoid layer. This result can already be used as a rough posterior estimate, however results can be further improved by combining with Importance Sampling. This is discussed in Sections~\ref{section:sampling_joint} and~\ref{section:hybrid} and is empirically verified in Section~\ref{sec:results}.

\section{Sequential UM for Importance Sampling}
\label{section:sampling_joint}
We now have a discriminative model which, given a set of observations $\mathbf{X_\mathcal{O}}$, will approximate all the posterior marginals. While useful on its own, the estimated marginals are not guaranteed to be unbiased. To obtain a guarantee of asymptotic unbiasedness, while making use of the speed of the approximate solution, we use the UM for proposals in importance sampling. A naive approach might be to sample each $X_i \in \XU$ independently from $\UM(\xxO)_i$, where $\UM(\xxO)_i$ is $i$-th element of the vector $\UM(\xxO)$.  However, the product of the (approximate) posterior marginals may be very different to the true posterior joint, even if the marginal approximations are good. A problematic example of this, where the variance of weights becomes very large, is highlighted in Appendix~\ref{subsection:problematic_example}.

The universality of the UM  makes possible a scheme we call \textit{Sequential Universal Marginalizer Importance Sampling} (SUM-IS).  A single proposal sample $\xS$ is generated sequentially as follows. First, introduce a new partially observed state $\xxSuO$ initialized to $\xxO$. Sample $[\xS]_1 \sim \UM(\xxO)_1$, and update $\xxSuO$ so that $X_1$ is now observed with this value.  Now we repeat the process, at each step sampling $[\xS]_i \sim \UM(\xxSuO)_i$, and updating $\xxSuO$ to include the new sampled value. Thus, we can approximate the conditional marginal for a node $i$ given the current sampled state $\mathbf{X_\mathcal{S}}$ and evidence $\mathbf{X_\mathcal{O}}$, to get the optimal proposal $Q_i^\star$ as: 
\begin{align}
Q^\star_i &= P(X_i|\{X_1,\ldots, X_{i-1}\}\cup\mathbf{X_\mathcal{O}})
\approx \UM(\xxSuO)_i = Q_i.
\end{align}
The full sample $\xS$ is thus drawn from an implicit encoding by the UM of the (approximate) posterior \emph{joint} distribution, as can be seen by observing the product of sample probabilities (Equation~\ref{eq:topo}), so may be expected to yield low variance importance weights when used as the proposal distribution. 
\begin{align}
  \label{eq:topo}
  Q &= \UM(\xxO)_1\prod_{i=2}^{N}\UM(\xxSuO)_i
  \approx P(X_1|\mathbf{X_\mathcal{O}})\prod_{i=2}^{N}P(X_i|X_1,\ldots, X_{i-1},\mathbf{X_\mathcal{O}}).
\end{align}
The process by which we sample from these approximately optimal proposals is illustrated in Algorithm~\ref{algorithm:sumis} and in Figure~\ref{fig:inf} in Appendix~\ref{section:process}. 
This procedure requires that nodes are sampled sequentially, using the UM to provide a conditional probability estimate at each step. This can affect computation time, depending on the parallelization scheme used for sampling. However, some parallelization efficiency can be recovered by increasing the number of samples, or batch size, for all steps. Alternatively a hybrid method which approximates the joint but requires only one call of the UM is proposed in Section~\ref{section:hybrid}.
\section{Hybrid Proposals}
\label{section:hybrid}
The full SUM-IS process requires sequential sampling and many evaluations of the UM, which may be costly.  We also explored a heuristic scheme by which a single UM output of all marginals may be combined with ancestral sampling, when nodes are sampled in topological order. 

The proposal distribution for each node $X_i$ is a mixture of the UM marginal $\UM(\xxO)_i$, and the conditional $P(X_i | \xSnPai)$, where $\xSnPai$ encodes the (sampled or evidential) observations of all ancestors of $X_i$.  Note that this conditional can be calculated directly from the graph. The scheme uses a mixture model, with
$$Q(X_i) = \beta\cdot \UM(\xxO)_i + (1 - \beta)\cdot P(X_i | \xSnPai).$$
Here, each node in the proposal is drawn either from the UM approximate marginal given the observed evidence, independently of previously sampled nodes, or according to its prior dependence on previously sampled nodes (and any ancestral evidence), independently of evidence nodes that fall later in the topological sequence.  This approach expects to blend these two forms of dependence,  generating a reasonable IS proposal - described in Algorithm~\ref{algorithm:hybrid} in the Appendix~\ref{section:algorithms}. We compare different fixed $\beta$  values in Section~\ref{sec:results}. However, $\beta$ can also be a function of the currently sampled state and the observations. This is left for future work.
\section{Methods}
\label{sec:methods}
We trial feed-forward NN architectures with a hyperparameter search on the number and size of hidden layers. The quality of conditional marginals is measured using a test set of posterior marginals computed for multiple sets of evidence via ancestral sampling with 300 million samples. Two main metrics are considered - overall mean absolute error (absolute difference between the true and predicted node posterior) and mean maximum absolute error (maximum absolute difference averaged across the evidence sets). For importance sampling results we also examine the Pearson correlation of the true and predicted marginal vectors, as well as Effective Sample Size (ESS). Kish's ESS is defined as $\big(\sum_{j=1}^{M} w_j\big)^2/\sum_{j=1}^{M} w_j^2$. The best performing UM is used for subsequent experiments using the hybrid proposals scheme proposed in Section~\ref{section:hybrid}.

We use ReLU non-linearities, apply a dropout with a probability of $0.5$ after each hidden layer and use the Adam optimization method~\citep{kingma2014adam}. We consider two encoding schemes for unobserved and observed nodes: \begin{enumerate*}

\item[] \textbf{2-bit Representation:} Two binary values representation. One binary value represents whether the node is observed, the other represents (if observed) whether it is True or False.
\item[] \textbf{33-bit Representation (1-bit + 32-bit Continuous):} One binary node represents whether the node is observed and another continuous node is in $\{0, 1\}$ if observed and the prior probability if not observed.
\end{enumerate*}

\section{Results}
\label{sec:results}
\noindent
{\bf  UM Architecture  and Representation Search.}
We run a hyperparameter search on network size and unobserved representation, reporting the results in Table~\ref{tab:table_2}. The largest one layer network performs the best. The difference between the representations is not large, but the results suggest that providing the priors may help improve performance.
\begin{table}[h]
  \centering
  \begin{tabular}{ccccc}
    \toprule
    Units per hidden layer & \multicolumn{2}{c}{2-bit} & \multicolumn{2}{c}{33-bit (priors)} \\
	&$|e|$&$\max(|e|)$&$|e|$&$\max(|e|)$\\
    \midrule
    (2048) & 0.0063 & 0.3425  & 0.0060 & 0.3223  \\
    (4096) & 0.0053 & 0.2982 & \textbf{0.0052} & \textbf{0.2951}  \\
    (1024, 1024) & 0.0081 & 0.4492 & 0.0083 & 0.4726 \\ 
    (2048, 2048) & 0.0071 & 0.4076 & 0.0071 & 0.4264 \\
    \bottomrule
  \end{tabular}
  \caption{Average mean absolute error ($|e|$) and average maximum absolute error ($\max(|e|)$) of the UM evaluated on the test set after training on different sized one- and two-layer NN architectures for 20,000 iterations. Best values are highlighted in bold.}
  \label{tab:table_2}
\end{table}

\noindent
{\bf Hybrid Importance Sampling using the UM.}
We assess the change in performance on the evidence sets with increasing number of samples. An increase in the maximum achieved correlation is observed, as well as higher ESS (Table~\ref{table:ess} in Appendix~\ref{section:additional_results}). Figure~\ref{fig:approx-cond} indicates standard IS ($\beta= 0 $) reaches $92\%$ correlation after $2$ million samples, whereas hybrid proposals with $\beta = 0.25$ exceed $95\%$ after only $250,000$ samples, ultimately achieving $96\%$ correlation in $2$ million samples. We achieve both a higher accuracy and a significant reduction in computational cost per inference.
\begin{figure}[h]
\centering 
\includegraphics[width=1\textwidth]{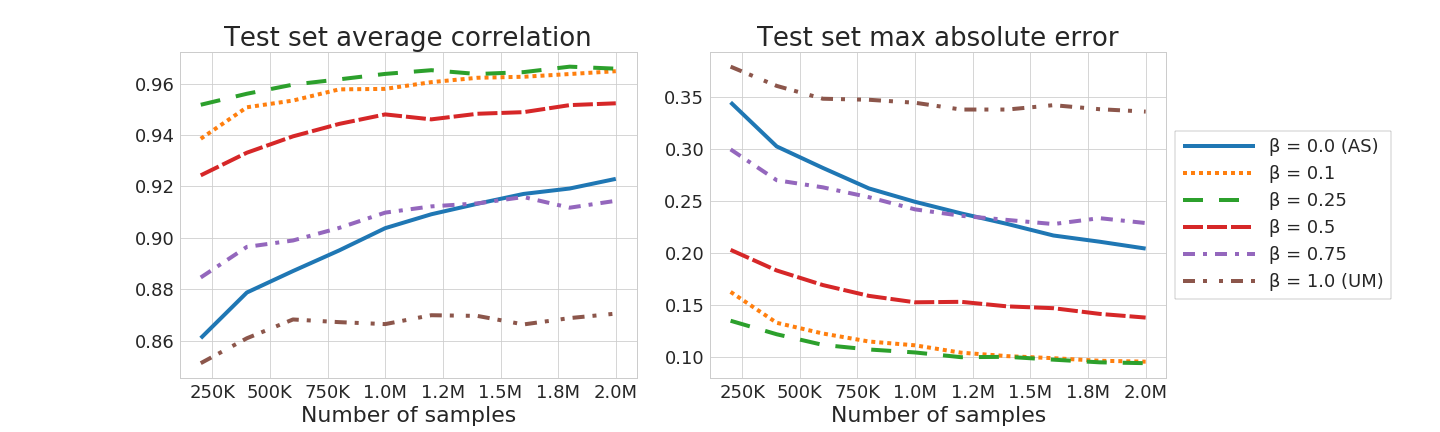}
\caption{
Hybrid importance sampling performance for various values of the mixing parameter $\beta$ between pure ancestral sampling proposals ($\beta = 0$) and UM marginals independent of the sampled state of unobserved nodes ($\beta = 1$). When $\beta \in [0.1, 0.5]$ we see better marginal estimates in 250k samples than obtained in all 2 million samples when not mixing in the predictions from the UM.}
\label{fig:approx-cond}
\end{figure}
\section{Conclusion}
This paper introduces a Universal Marginalizer, a neural network which can approximate all conditional marginal distributions of a BN. We have shown that a UM can be used via a chain decomposition of the BN to estimate the joint posterior and thus the optimal proposal distribution for importance sampling. While this process is more computationally intensive, a first-order approximation can be used requiring only a single evaluation of a UM per set of evidence. Our experiments show that the hybrid IS procedure delivers significant improvements in sampling efficiency.

\bibliographystyle{plainnat} \bibliography{refs}

\appendix 

\section{Importance Sampling}
\label{sec:app:impo-sampling}

We may use Importance Sampling (IS) to provide posterior marginal estimates, $P(\mathbf{X_\mathcal{U}}|\mathbf{X_\mathcal{O}})$ in BN inference. To do so, we draw samples $\xU$ from a distribution $Q(\XU|\XO)$, known as the \textit{proposal} distribution, which we can both sample and evaluate efficiently. Then, assuming that the numerator can be evaluated (which requires that $\XU$ contain the Markov boundary of $\XO$ along with all its ancestors), we have:
  \begin{equation}
    \label{eq:importance-sampling}
    \begin{aligned}
      P(\mathbf{X_\mathcal{U}=\xU}|\mathbf{X_\mathcal{O}}=\xO)
      &=\frac{Q(\xO)}{P(\xO)}\int 1_{\xU}(\mathbf x) \frac{P(\mathbf x,\xO)}{Q(\mathbf x, \xO)}Q(\mathbf x| \xO)d \mathbf x\\
      &=\lim_{n\to\infty} \sum_{i=1}^{n} 1_{\xU}(\mathbf x_i) \frac{w_i}{\sum_{j=1}^{n} w_j},\
    \end{aligned}
  \end{equation}

where $\mathbf x_i\sim Q$ and $w_i=P(\mathbf x_i, \xO)/Q(\mathbf x_i, \xO)$ are the \textit{importance sampling weights} and $1_{\xU}(\mathbf x)$ is an indicator function for $\xU$. 

The simplest proposal distribution is the prior, $P(\XU)$.  However, as the prior and the posterior may be very different this is often an inefficient approach. An alternative is to use an estimate of the posterior distribution as a proposal. This is the approach we develop.

\section{Sampling from the Posterior Marginals: A Problematic Example}
\label{subsection:problematic_example}
Take a BN with Bernoulli nodes and of arbitrary size and shape. Consider 2
specific nodes, $X_i$ and $X_j$, such that $X_j$ is caused only and always by
$X_i$:
\begin{align*}
  P(X_j=1|X_i=1) &= 1,\\
  P(X_j=1|X_i=0) &= 0.
\end{align*}
Given evidence $E$, we assume that $P(X_i|E)=0.001=P(X_j|E)$. We will now
illustrate that using the posterior distribution $P(X|E)$ as a proposal will not
necessarily yield the best result.

Say we have been given evidence, $E$, and the true conditional probability of $P(X_i|E) = 0.001$, therefore also $P(X_j|E) = 0.001$. We naively would expect $P(X|E)$ to be the optimal proposal distribution. However we can illustrate
the problems here by sampling with $Q=P(X|E)$ as the proposal.

Each node k $\in$ N will have a weight $w_k = P(X_k)/Q(X_k)$ and the total
weight of the sample will be $$w = \prod_{k=0}^{N} w_k.$$ The weights should be approximately 1 if Q is close to P. However, consider the $w_j$. There are four combinations of $X_i$ and $X_j$. We will sample $X_i$=1, $X_j$=1 only, in
expectation, one every million samples, however when we do the weight $w_j$ will be $w_j = P(X_j=1)/Q(X_j=1) = 1/0.001=1000$. This is not a problem in the limit, however if it happens for example in the first 1000 samples then it will
outweight all other samples so far. As soon as we have a network with many nodes whose conditional probabilities are much greater than their marginal proposals this becomes almost inevitable. A further consequence of these high weights is that, since the entire sample is weighted by the same weight, every node probability will be effected by this high variance.

\section{Process Diagrams}
\label{section:process}

\begin{figure}[h]
  \centering
  \begin{tikzpicture}[node distance=2cm, scale=0.8, transform shape]
    \node (PGM) [io] {1. Samples from PGM: $\textbf{X}\in \{0,1\}^N$.}; 
    \node (mask) [process, below of=PGM] {2. Mask $\textbf{X}$: $M:\{0,1\}\to \{0,1,*\}$,
      $M:$ probabilistic and/or deterministic.}; 
    \node (masked) [io, below of=mask] {3. Input: $M(\textbf{X})\in\{0,1,*\}^N$, Labels: \textbf{X}}; 
    \draw [arrow] (PGM) -- (mask);  
    \node (train) [process, below of=masked] {4. Neural network with sigmoid output.};
        \draw [arrow] (mask) -- (masked);  
          \node (output) [io, below of=train] {5. Output: Predicted
      Posterior $P(\textbf{X} | M(\textbf{X}))$}; 
      \draw [arrow] (train) -- (output); \draw [arrow] (masked) -- (train);
    \path [->, red, very thick] (train) edge[loop right] node (CE)
    [right, black] {Cross-Entropy loss training.} (train) ;
  \end{tikzpicture}
  \caption{The process to train a Universal Marginalizer using binary data generated from a Bayesian Network}
  \label{fig:train}
\end{figure}
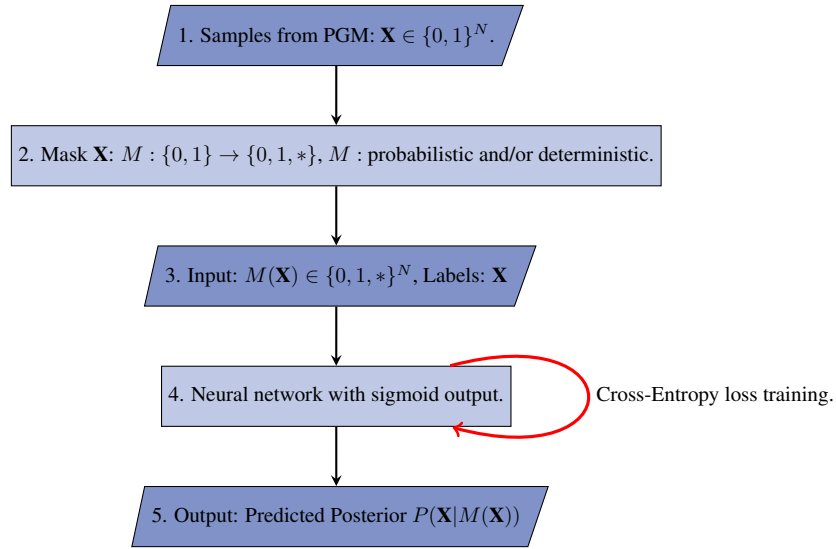

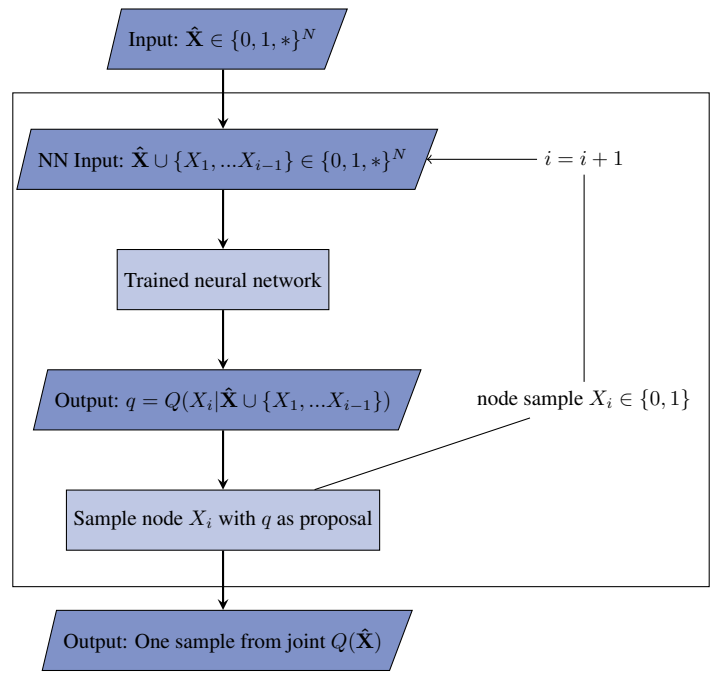
\begin{figure}[h]
  \centering
  \begin{tikzpicture}[node distance=2cm, scale=0.8, transform shape]
    \node (inp) [io] {Input: $\mathbf{\hat{X}}\in\{0,1,*\}^N$};
    \node (test_data) [io, below of=inp] {NN Input: $\mathbf{\hat{X}} \cup \{X_1,...X_{i-1}\} \in\{0,1,*\}^N$};
    \draw [arrow] (inp) -- (test_data);
    \node (train_output) [process, below of=test_data] {Trained neural network}; 
    \draw [arrow] (test_data) -- (train_output); 
    \node (sample) [io, below of=train_output] {Output: $q =
      Q(X_i | \mathbf{\hat{X}} \cup \{X_1,...X_{i-1}\})$}; \node (impo) [process, below
    of=sample] {Sample node $X_i$ with $q$ as proposal}; \draw
    [arrow] (sample) -- (impo); 
    \draw [arrow] (train_output) -- (sample); 
    \node (i) [right of=test_data,xshift=4cm] {$i = i+1$}; 
    \node (ce) [right of=sample,xshift=4cm] {node sample $X_i \in \{0,1\}$}; \draw [->, black] (impo) -- (ce) -- (i) -- (test_data) ; 
    \node (full_sample) [io, below of=impo] {Output: One sample from joint $Q(\mathbf{\hat{X}})$}; \draw [arrow] (impo) -- (full_sample); 
      \node[draw,inner sep=13mm, fit=(test_data) (train_output) (sample) (i) (ce)(impo)] {};
  \end{tikzpicture}
  \caption{Importance Sampling + UM: The part in the box is repeated $N$ times, for each node $i$ in topological order}
  \label{fig:inf}
\end{figure}

\newpage
\section{Algorithms}
\label{section:algorithms}

\begin{algorithm}
\caption{Sequential Universal Marginalizer importance sampling}
\label{algorithm:sumis}
\begin{algorithmic}[1]
\State Order the nodes topologically $X_1,...X_N$, where $N$ is the total number of nodes.
\For{$j$ in [1,...,$M$] (where $M$ is the total number of samples):}
  \State $\xxS = \emptyset$
  \For{$i$ in [1,...$N$]:}
      \State sample node $x_i$ from $Q(X_i)=\UM(\xxSuO)_i \approx P(X_i |\mathbf{X_\mathcal{S}}, \mathbf{X_\mathcal{O}})$
      \State add $x_i$ to $\xxS$
  \EndFor
  \State $[\xS]_j = \xxS$
  \State $w_j = \prod_{i=1}^{N}\frac{P_i}{Q_i}$ (where $P_i$ is the likelihood, $P_i = P(X_i=x_i| \xSnPai)$ and $Q_i=Q(X_i=x_i)$)
\EndFor
\State $E_p[X] = \frac{\sum_{j=1}^{M}X_j w_j}{\sum_{j=1}^{M}w_j}$ (as in standard IS)
\end{algorithmic}
\end{algorithm}

\begin{algorithm}
\caption{Hybrid UM-IS}
\label{algorithm:hybrid}
\begin{algorithmic}[1]
\State Order the nodes topologically $X_1,...X_N$, where $N$ is the total number of nodes.
\For{$j$ in [1,...,$M$] (where $M$ is the total number of samples):}
  \State $\xxS = \emptyset$
  \For{$i$ in [1,...$N$]:}
      \State sample node $x_i$ from $Q(X_i) = \beta \UM(\xxO)_i + (1 - \beta) P(X_i=x_i | \xSnPai)$
      \State add $x_i$ to $\xxS$
  \EndFor
  \State $[\xS]_j = \xxS$
  \State $w_j = \prod_{i=1}^{N}\frac{P_i}{Q_i}$ (where $P_i$ is the likelihood, $P_i = P(X_i=x_i| \xSnPai)$ and $Q_i=Q(X_i=x_i)$)
\EndFor
\State $E_p[X] = \frac{\sum_{j=1}^{M}X_j w_j}{\sum_{j=1}^{M}w_j}$ (as in standard IS)
\end{algorithmic}
\end{algorithm}

\newpage
\section{Additional Results}
\label{section:additional_results}

\begin{table}[h]
  \caption{Effective sample size for hybrid UM-IS scheme with 2 million samples}
  \label{table:ess}
  \centering
  \begin{tabular}{cllllll}
    \toprule
     $\boldsymbol{\beta}$ & $0.0$ & $0.1$ & $0.25$ & $0.5$ & $0.75$ & $1.0$ \\
    \midrule
    \textbf{ESS} & 7678 & 15458 & 11779 & 1218 & 171 & 92 \\
    \bottomrule
  \end{tabular}
\end{table}

\end{document}